\documentclass[twoside,11pt]{article}

\usepackage{jmlr2e}
\usepackage{lastpage}
\usepackage{booktabs}
\usepackage{amsmath}
\usepackage[safe]{tipa}
\usepackage{xcolor}
\usepackage{xurl}
\usepackage{adjustbox}

\newcommand{\lmp}[1]{\textit{#1}}
\newcommand{\gl}[1]{`#1'}
\newcommand{\chrfT}{chrF++$_T$}
\newcommand{\chrfS}{chrF++$_S$}
\newcommand{\eps}{\textepsilon}
\newcommand{\ubar}{\textbaru}
\newcommand{\eng}{\textipa{N}}
\newcommand{\pmsd}[1]{{\scriptsize$\pm$#1}}

\jmlrheading{}{2026}{1-\pageref{LastPage}}{}{}{}{Samuel Babalola, Oche David Ankeli and Samiratu Ntohsi}
\makeatletter
\def\ps@jmlrtps{\let\@mkboth\@gobbletwo
  \def\@oddhead{}\def\@evenhead{}\def\@evenfoot{}%
  \def\@oddfoot{\parbox[t]{\textwidth}{\raggedright\scriptsize
    Proceedings of the 2026 AI for African Languages Conference. PMLR 2026.\\[3pt]
    \copyright\ 2026 S.~Babalola, O.D.~Ankeli \& S.~Ntohsi. License: CC-BY 4.0,
    see \url{https://creativecommons.org/licenses/by/4.0/}.\hfill}}}
\makeatother

\ShortHeadings{Tone Generation in an Unsupported Language}{Babalola, Ankeli and Ntohsi}
\firstpageno{1}

\begin{document}

\title{Tone Generation in an Unsupported Language: Evaluating\\ English-to-Limbum Translation with NLLB-200}

\author{\name Samuel Babalola \email sbabalola@alueducation.com \\
       \addr Ruzivo Research Lab, African Leadership University, Kigali, Rwanda\\
       Wavira AI\\
       ML Collective
       \AND
       \name Oche David Ankeli \email o.ankeli@alustudent.com \\
       \addr Ruzivo Research Lab, African Leadership University, Kigali, Rwanda\\
       Wavira AI
       \AND
       \name Samiratu Ntohsi \email sntohsi@alueducation.com \\
       \addr Ruzivo Research Lab, African Leadership University, Kigali, Rwanda}

\makeatletter\def\@starteditor{}\def\@editor{}\def\@endeditor{}\makeatother

\maketitle

\begin{abstract}%
Many African languages write tone with diacritics, so a single wrong mark can change a word's
meaning while the sentence still reads fluently. We ask whether a pretrained translation model
can learn to write tone in a language it never saw during pretraining, by fine-tuning NLLB-200 on
about 7{,}000 English--Limbum sentence pairs (Limbum is a Grassfields Bantu language of Cameroon).
We first find that most of the corpus is stored in a legacy font encoding that disguises vowel
letters and tone marks as French accented characters; we convert and verify it. In the
converted text, 41\% of words carry a tone mark and 35\% of running words are \emph{contested}:
the same word appears with different tone marks depending on context. When the model translates a
word correctly, it marks tone correctly 98.5\% of the time for words with a fixed tone pattern,
but only 88.1\% of the time for contested words. Overall translation quality is modest (chrF++
44.3), so tone can be scored on only about half of the reference words, those the model translates
correctly. Removing every tone mark changes chrF++ by just
1.9 points, so standard metrics barely register these errors. Neither encoding tone marks as
separate tokens nor restoring tone after toneless translation improves on direct fine-tuning, and
restoration is significantly worse for tone. A native speaker confirmed that most real tone errors
change the meaning, but only about one in five automatic tone mismatches is a genuine error. The legacy encoding barely affects the model but would have invalidated every tone analysis.
We release the converted corpus, code and model outputs.
\end{abstract}


\section{Introduction}
\begin{figure}[t]
\centering
\includegraphics[width=\linewidth]{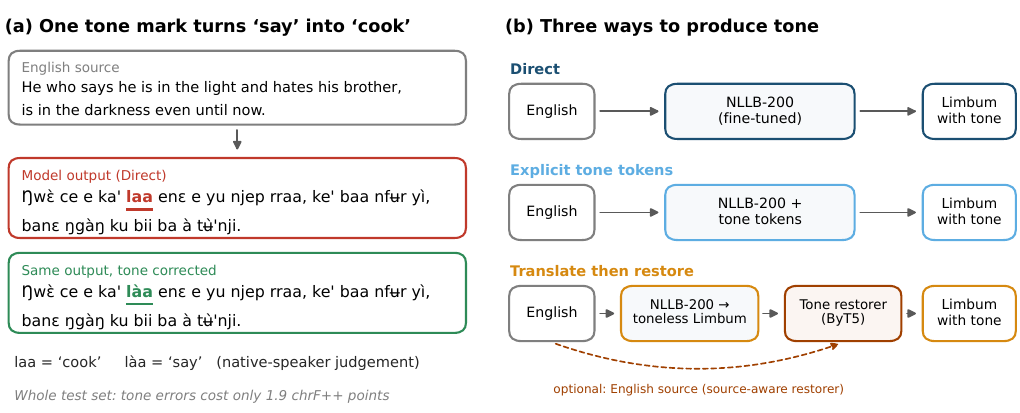}
\caption{(a) A real Direct output (top) and the same output with its one tone error corrected
(bottom). Writing \lmp{laa} \gl{cook} instead of \lmp{l\`aa} \gl{say} changes the meaning, as
judged by our native-speaker author, yet over the whole test set tone errors move chrF++ by under
two points. (b) The three ways of producing tone that we compare.}
\label{fig:overview}
\end{figure}

Pretrained multilingual translation models such as NLLB-200 \citep{nllb2022} have made it
practical to build a first translation system for a new language from a few thousand
sentence pairs \citep{adelani2022few}. Most of this evidence concerns translation
\emph{out of} the new language. Generation \emph{into} languages that are absent or weakly
represented in pretraining is harder: \citet{omt2026} report that models which can
\emph{interpret} undersupported languages often cannot \emph{generate} them with
meaningful fidelity.

For tonal languages written with tone diacritics, generation has an additional failure
mode. Limbum distinguishes words by pitch alone, for example \lmp{b\`a} \gl{bag},
\lmp{b\=a} \gl{fufu} and \lmp{b\'a} \gl{hill} \citep{gjersoe2019hybrid}, and tone also
carries grammatical information. An output that places the wrong mark can be fluent and still
wrong (Figure~\ref{fig:overview}a). Character-level metrics such as chrF++ \citep{popovic2017chrf} give a word with a
wrong mark most of its credit, so this error is largely invisible in the numbers usually
reported.

In earlier work we fine-tuned NLLB-200 for Limbum$\rightarrow$English
\citep{ntohsi2026limbum}. In that direction tone marks appear only in the source, so it
remains open whether a pretrained model can learn to \emph{write} Limbum tone. We study the
English$\rightarrow$Limbum direction and ask whether the corpus's tone orthography is
consistent and representable by the model's tokenizer, how accurately a fine-tuned NLLB-200
writes tone in a language absent from its pretraining, whether representing tone marks as
separate tokens or restoring tone after translation improves on direct generation
(Figure~\ref{fig:overview}b), and where the remaining errors concentrate.

Examining the orthography uncovered a problem that affects every other question. The New
Testament, which makes up 84\% of the corpus, is stored in a legacy 8-bit font encoding in
which vowel letters look like tone marks and tone marks look like letters. We convert and
verify the text, and then measure what the encoding does to training and to evaluation.

On the converted data, direct fine-tuning writes the correct tone on 98.5\% of correctly
translated words that have a fixed tone pattern, but only on 88.1\% of \emph{contested}
words, whose tone pattern varies with context and which make up 35\% of running text.
Removing every tone mark changes chrF++ by just 1.9 points, so this gap is nearly invisible
to the metric. Neither explicit tone tokens nor translate-then-restore pipelines close it,
and restoration is significantly worse for tone in every seed. The legacy encoding, in
contrast, barely affects the models but would have invalidated every tone analysis.

Our contributions are fourfold. We provide the first measurement of tone-generation accuracy
for a language unseen by a pretrained MT model, separating fixed-pattern from contested
words. We compare direct generation, explicit tone tokens and three translate-then-restore
pipelines under identical conditions, with three seeds and paired significance tests. We
propose a tone-aware evaluation protocol that reports word-level tone accuracy together with
coverage and an oracle restoration ceiling. Finally, we release\footnote{\url{https://github.com/SammyGbabs/limbum-tone-mt}} a verified Unicode conversion
of the Limbum New Testament and show how a legacy encoding distorts analysis and scores, a
practical warning for the many Bible-derived corpora of African languages.

\section{Background and Related Work}

\paragraph{Limbum tone.}
Limbum is a Grassfields Bantu language spoken mainly in the Donga-Mantung division of
North-West Cameroon, with estimates of its speaker population ranging from about 130{,}000
\citep{omniglotlimbum} to 1.3 million \citep{gjersoe2019hybrid}. Descriptive work analyses it as
having three level tones (high, mid, low) and four contour tones \citep{fiore1987,fransen1995},
and tone interacts with phrase position: low-falling contours surface only phrase-finally
\citep{gjersoe2019hybrid}. Linguistic transcriptions mark all three levels, with a macron for
mid tone \citep{gjersoe2019hybrid}. The practical orthography of our corpus and of published
Unicode Limbum text \citep{omniglotlimbum} marks low tone with a grave accent, falling tone
with a circumflex and rising tone with a caron, and leaves high and mid tone unmarked
(\S\ref{sec:audit}).

\paragraph{Tone diacritics in African-language NLP.}
Most computational work on tone marking targets Yor\`ub\'a, which also has three level tones
written with acute, grave and an optional mid mark. Diacritic restoration has been modelled
as sequence-to-sequence learning \citep{orife2018attentive,orife2020improving}, benchmarked
with T5-family models in YAD \citep{olawole2024yad}, and studied for other languages with
character- and byte-level models \citep{naplava2018diacritics,stankevicius2022byt5}.
MENYO-20k \citep{adelani2021menyo} analysed how diacritics in the \emph{training data} affect
Yor\`ub\'a--English MT. All of this work concerns a language that is \emph{seen} in
pretraining and has tens of thousands of curated sentences or more. Our setting differs in
three ways: the target is unseen by the translation model, the data is about an order of
magnitude smaller, and we evaluate tone in the generated output of end-to-end MT systems as
well as in restoration of gold text.

\paragraph{Adapting multilingual MT to unseen languages.}
Adapting a pretrained multilingual model to a new language usually means introducing or
repurposing a language tag \citep{johnson2017google} and fine-tuning on a small corpus
\citep{neubig2018rapid,adelani2022few}. For NLLB-200, practitioners commonly initialize a new
tag from a related language's tag \citep{dale2023nllb}. For Limbum$\rightarrow$English,
our earlier work found that multilingual transfer dominates \citep{ntohsi2026limbum} (chrF++ 47.3 versus 14.5
for a Transformer trained from scratch). We hold tag initialization fixed and focus on the
generation direction.

\section{Data}
\label{sec:data}

\subsection{Corpus and split}
We use the Limbum--English corpus released in our earlier work \citep{ntohsi2026limbum}: New Testament verses aligned
with English, and entries from a bilingual dictionary. After de-duplication and the
exclusions below, 8{,}829 pairs remain. Random verse-level splits put adjacent verses into
both training and test, so we split the New Testament into 20 contiguous blocks of verses
(16 train, 2 development, 2 test) and the dictionary randomly (80/10/10). This yields
7{,}062 training, 884 development and 883 test pairs (738 verses and 145 dictionary entries).
Our numbers are therefore not comparable to the random split of the earlier work.

\subsection{Recovering the orthography from a legacy encoding}
\label{sec:encoding}
The New Testament and the dictionary use disjoint character inventories. The New Testament
contains no \eps, \ubar\ or \eng\ at all, but 33{,}114 \lmp{\'a}, 18{,}842 \lmp{\~n} and
12{,}137 \lmp{\'u}; the dictionary contains 1{,}479 \eps, 869 \eng\ and 564 \ubar\ and none
of those accented letters. In the New Testament, \lmp{\`e} follows a vowel in all but 22 of
its 84{,}487 occurrences, so it is not a letter but a tone glyph typed after the vowel it
belongs to. This is the signature of a legacy 8-bit font in which special letters and
combining diacritics were mapped onto French accented characters.
Table~\ref{tab:mapping} gives the mapping we derived. Applied to Luke~11:2, it reproduces
the Unicode text published by \citet{omniglotlimbum} letter for letter and tone for tone; for
example \lmp{E m la\`ea ne\`e wowe\`ee en\'a\^e} becomes
\lmp{E m l\`aa n\`e wow\`ee en\^{\eps}}.

\begin{table}[t]
\centering\small
\caption{Legacy font characters in the New Testament text and their Unicode values. Tone
glyphs follow the vowel they modify; wide and narrow variants exist.}
\label{tab:mapping}
\begin{tabular}{lll}
\toprule
Legacy & Unicode & Role \\
\midrule
\'a, \'A & \eps\ (and capital) & vowel letter \\
\'u & \ubar & vowel letter \\
\~n, \~N & \eng\ (and capital) & consonant letter \\
\"i & i & base for a following tone glyph \\
\`e, \`\i\ (after a vowel) & combining grave (U+0300) & low tone \\
\^e, \^a, \^\i, \^u (after a vowel) & combining circumflex (U+0302) & falling tone \\
\bottomrule
\end{tabular}
\end{table}

The conversion attached 102{,}442 tone glyphs to their vowels and left 13 unattached. All 13
come from 13 training verses that mix legacy glyphs with genuine Unicode accented letters;
these cannot be converted reliably and were excluded. We also repaired 43 rows (29 dictionary,
14 New Testament) in which a tone mark was duplicated or separated from a syllabic nasal by a
space, and replaced curly quotation marks, which NLLB-200's tokenizer maps to the unknown
token, with ASCII quotes.

\subsection{Tone in the converted corpus}
\label{sec:audit}

\begin{table}[t]
\centering\small
\caption{Corpus and tokenizer audit after conversion (training split). Values for the
unconverted legacy text are given in parentheses where they differ.}
\label{tab:audit}
\begin{tabular}{lr}
\toprule
Statistic & Value \\
\midrule
Words with $\geq$1 tone mark, New Testament / dictionary & 40.6\% / 44.3\% \\
Tone marks: grave / circumflex / caron & 96{,}979 / $\approx$9.8k / 118 \\
Acute and macron & 0 \\
Contested types, all / frequency $\geq$5 & 10.0\% / 24.4\% \\
Running words that are contested & 35.4\% \\
\midrule
NLLB round trip exact & 100.0\% (70.0\%) \\
Tone-mark survival in round trip & 100.0\% \\
\texttt{<unk>} per 100 tokens & 0.00 (0.95) \\
Tokens per word: toned / toneless / explicit & 1.86 / 1.59 / 2.16 \\
& (2.05 / 1.76 / 2.53) \\
\bottomrule
\end{tabular}
\end{table}

Table~\ref{tab:audit} summarizes the converted corpus. About 41\% of words carry a tone
mark; almost all are grave (low tone), followed by circumflex (falling) and caron (rising,
dictionary only). No acute or macron occurs, so high and mid tone are unmarked in both
sources.

We call a word type \emph{contested} if, in the training targets, its second most frequent
toned spelling accounts for at least 10\% of its occurrences and occurs at least three
times. Contested types are 24.4\% of types seen at least five times but 35.4\% of running
words, because they include very frequent grammatical words: \lmp{a}/\lmp{\`a} (11{,}736
vs.\ 9{,}351 occurrences), \lmp{yi}/\lmp{y\`\i}, \lmp{laa}/\lmp{l\`aa},
\lmp{gee}/\lmp{g\`ee}, \lmp{du}/\lmp{d\`u}, \lmp{en\eps}/\lmp{en\^{\eps}}. These are the
words where tone must be inferred from context.

NLLB-200's tokenizer \citep{kudo2018sentencepiece} knows \eps, \eng\ and \ubar: after
conversion every training sentence survives an encode--decode round trip exactly, with no
unknown tokens and every tone mark intact. Before conversion, 30\% of sentences failed the
round trip because of curly quotes, and the legacy spelling needed 10\% more tokens per word.

\section{Methodology}
\label{sec:method}

\subsection{Translation systems}
\label{sec:systems}

All systems fine-tune NLLB-200-distilled-600M with the same data, adaptation recipe and
decoding; they differ only in how the target side represents tone. We study NLLB-200 because it
is a widely used open translation model covering many African languages and is commonly adapted
to new African languages in practice \citep{adelani2022few,dale2023nllb}; it was also the basis of
our earlier Limbum work, and Limbum is absent from it, which makes it a clean test of an unseen
language. The distilled 600M variant fits a single 16\,GB GPU, which allowed three seeds for every
system.

\paragraph{Shared adaptation.}
We add a target tag \texttt{lmp\_Latn} initialized as the mean of the Swahili, Luganda and
Lingala tags, as in our earlier work \citep{ntohsi2026limbum}; it is the decoder's forced first token.

\paragraph{Direct.} Trained on toned Limbum targets with the unmodified tokenizer.

\paragraph{Explicit tone tokens.} Targets are decomposed (NFD) and each tone mark (grave,
circumflex, caron) is mapped to a dedicated new vocabulary item placed directly after its
base letter; the tone-free text is tokenized as usual. Every tone decision becomes a separate
prediction. New tone embeddings are initialized from the pretrained embedding of the
corresponding combining character.

\paragraph{Translate then restore.} NLLB-200 is trained on toneless targets, and a separate
restorer adds tone: (i) \emph{lexical}, the most frequent toned spelling of each word in the
training targets; (ii) \emph{ByT5}, ByT5-small \citep{xue2022byt5} fine-tuned to map toneless
chunks of up to 30 words to toned chunks; (iii) \emph{source-aware ByT5}, which also receives
the English source sentence. A consistency guard restricts restorers to adding tone: any
output word whose toneless form differs from the input word is replaced by the input word
(this never triggered on gold input).

\subsection{Experimental setup}
\label{sec:setup}
NLLB systems are trained with Adafactor \citep{shazeer2018adafactor} (learning rate
$10^{-4}$, 200 warm-up steps, linear decay), effective batch size 16, fp16, for up to 8
epochs with early stopping (patience 2) on development chrF++. No run stopped early;
development scores were still rising by about 0.2 points per epoch at epoch 8 for all
systems, so absolute scores could improve with longer training, but all systems received the
same budget. Test decoding uses beam search (beam 4). ByT5 restorers use Adafactor (learning
rate $5\times10^{-4}$), effective batch size 16, fp32, up to 6 epochs with early stopping on
development TAcc. Everything runs on a single 16\,GB T4 GPU; settings are listed in
Appendix~\ref{app:hparams}.

\section{Evaluation}
\label{sec:eval}

\paragraph{Toned and toneless chrF++.}
We compute chrF++ \citep{popovic2017chrf} with sacreBLEU \citep{post2018call} on toned text
(\chrfT) and after removing tone marks from hypothesis and reference (\chrfS). The gap
$\Delta=\text{\chrfS}-\text{\chrfT}$ is the score lost to tone alone.

\paragraph{Word-level tone accuracy and coverage.}
We align hypothesis and reference words by their toneless forms (longest matching blocks).
\textbf{TAcc} is the share of aligned words whose toned forms match exactly;
\textbf{TAcc$_M$} restricts this to reference words carrying at least one mark, so that
leaving words unmarked is not rewarded. Tone can only be scored on words the system
translated correctly, so we always report \textbf{coverage}, the share of reference words
aligned. The \textbf{mark ratio} (MR) is the number of hypothesis marks divided by reference
marks. We break TAcc down into contested and unambiguous words, and phrase-final versus
phrase-medial position.

\paragraph{Statistics.}
NLLB systems are trained with three seeds; we report mean $\pm$ standard deviation.
Restorers are trained once and applied to each seed's toneless output. Each system is
compared with Direct by paired bootstrap resampling over test sentences
\citep{koehn2004statistical} (1{,}000 resamples, per seed).

\paragraph{Legacy versus Unicode.}
To isolate the effect of the encoding, we take models trained on the unconverted legacy text
(same split, same recipe), convert their outputs with the same mapping, and score them
against the same Unicode references.

\paragraph{Human evaluation.}
A native Limbum speaker among the authors judged 100 test sentences containing contested words,
comparing Direct with source-aware ByT5 restoration. Outputs were shown in random A/B order
without system labels, and the annotator did not see the assignment key. For each output, the annotator
counted tone errors and marked whether any tone error changes the meaning, then gave an overall
preference. Results are reported in \S\ref{sec:human}.

\section{Results}

\begin{table*}[t]
\centering\small
\caption{English$\rightarrow$Limbum test results, mean $\pm$ s.d.\ over three NLLB seeds.
\chrfT/\chrfS: chrF++ on toned / toneless text; $\Delta$: score lost to tone.
TAcc$_M$: tone accuracy on marked reference words; TAcc$_{con}$: on contested words.
MR: mark ratio; Cov.: share of reference words aligned (the words tone is scored on).
$\downarrow$: tone accuracy significantly lower than Direct ($p<0.01$) in every seed.}
\label{tab:main}
\setlength{\tabcolsep}{4pt}
\begin{adjustbox}{max width=\linewidth}
\begin{tabular}{lrrrrrrrrr}
\toprule
System & \chrfT & \chrfS & $\Delta$ & BLEU & TAcc & TAcc$_M$ & TAcc$_{con}$ & MR & Cov. \\
\midrule
Direct & 44.3\pmsd{0.1} & 46.2\pmsd{0.1} & 1.9 & 26.6 & 94.7\pmsd{0.1} & 92.9\pmsd{0.1} & 88.1\pmsd{0.2} & 1.03 & 49.6 \\
Explicit tone tokens & 44.4\pmsd{0.1} & 46.3\pmsd{0.1} & 1.9 & 26.7 & 94.6\pmsd{0.2} & 93.1\pmsd{0.4} & 87.7\pmsd{0.6} & 1.09 & 49.5 \\
\midrule
Toneless, no restoration & 30.0\pmsd{0.1} & 47.0\pmsd{0.3} & 17.0 & 9.6 & 60.1\pmsd{0.1} & 0.0 & 53.2\pmsd{0.2} & 0.00 & 50.4 \\
\quad + lexical restore & 41.1\pmsd{0.2} & 47.0\pmsd{0.3} & 5.9 & 20.7 & 86.0\pmsd{0.1}$\downarrow$ & 71.0\pmsd{0.2} & 64.2\pmsd{0.2} & 0.83 & 50.4 \\
\quad + ByT5 restore & 44.9\pmsd{0.3} & 47.0\pmsd{0.3} & 2.1 & 26.9 & 93.9\pmsd{0.2}$\downarrow$ & 91.9\pmsd{0.2} & 86.0\pmsd{0.4} & 0.99 & 50.4 \\
\quad + source-aware ByT5 & 44.9\pmsd{0.3} & 47.0\pmsd{0.3} & 2.1 & 27.0 & 94.0\pmsd{0.0}$\downarrow$ & 91.7\pmsd{0.1} & 86.1\pmsd{0.2} & 0.98 & 50.4 \\
\bottomrule
\end{tabular}
\end{adjustbox}
\end{table*}

\subsection{Direct fine-tuning learns fixed tone but not contested tone}
Table~\ref{tab:main} gives the main results. Because the tokenizer preserves every tone mark
(\S\ref{sec:audit}), differences between systems reflect what they learn, not what they can
represent. Direct fine-tuning reaches \chrfT\ 44.3 and
writes the correct tone on 94.7\% of aligned words (92.9\% of tone-marked words). This
average hides a sharp split (Table~\ref{tab:breakdown}): 98.5\% on words with a fixed tone
pattern but 88.1\% on contested words, so roughly one contested word in eight is written
with the wrong tone. Since contested words are 35\% of running text, these errors are
frequent, yet removing every tone mark from output and reference raises chrF++ by only
$\Delta=1.9$ points. A character-level score alone cannot tell whether a tonal language is
being written correctly. The mark ratio of 1.03 shows that the model neither omits nor
over-produces tone; it misplaces it.

\subsection{Alternative tone pipelines do not help}
\begin{figure}[t]
\centering
\includegraphics[width=\linewidth]{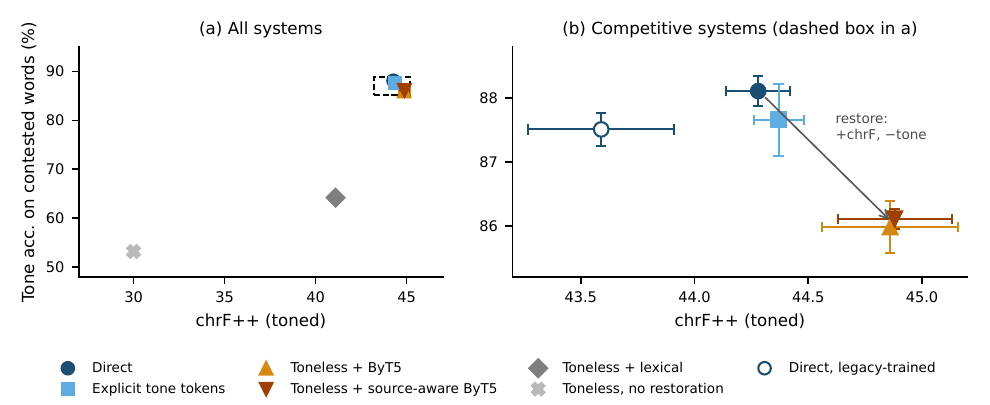}
\caption{Overall quality versus tone accuracy on contested words (mean $\pm$ s.d.\ over three
seeds). (a) Removing tone, or restoring it lexically, costs both. (b) Among competitive systems,
chrF++ varies by about 1 point while tone accuracy varies by 2: restoration moves right (small,
unreliable chrF++ gain) and down (reliable tone loss). The hollow point is Direct trained on the
legacy encoding, scored on Unicode references.}
\label{fig:tradeoff}
\end{figure}

\emph{Explicit tone tokens} change nothing: \chrfT\ 44.4 vs.\ 44.3 and TAcc 94.6 vs.\ 94.7,
with no significant difference in either direction in any seed. This is expected: the tokenizer already preserves every mark, so there is nothing for explicit tokens to
repair.

\emph{Translate then restore} is worse for tone (Figure~\ref{fig:tradeoff}). Without restoration, toneless output loses
14 \chrfT\ points, confirming that tone carries real weight in the metric when it is absent
altogether. Lexical restoration recovers fixed-pattern words (98.2\%) but fails on contested
words (64.2\%). Both ByT5 restorers reach \chrfT\ 44.9, higher than Direct by 0.3--1.1
points, but this gain is significant in only one of three seeds. Their tone accuracy is
lower than Direct's by 0.5--1.0 points (TAcc), and this difference is significant in every
seed ($p<0.01$). Giving the restorer the English source changes little (86.1\% vs.\ 86.0\%
on contested words).

Table~\ref{tab:oracle} explains why. Restoring tone on the \emph{gold} toneless reference,
the best case for any restorer, reaches 91.3--91.7\% on contested words. Direct generation
(88.1\%) is within a few points of this ceiling, and the pipeline loses further accuracy
because restoration operates on imperfect translations. Part of the tone information in
contested words is not recoverable from the written Limbum context alone.

\begin{table}[t]
\centering\small
\caption{Restorers applied to the gold toneless test references (an upper bound for
translate-then-restore). TAcc$_{con}$: contested words.}
\label{tab:oracle}
\begin{tabular}{lrrrr}
\toprule
Restorer & TAcc & TAcc$_M$ & TAcc$_{con}$ & \chrfT \\
\midrule
Lexical (most frequent) & 85.7 & 71.3 & 65.1 & 83.7 \\
ByT5 & 95.6 & 93.4 & 91.3 & 94.9 \\
Source-aware ByT5 & 95.8 & 93.4 & 91.7 & 95.1 \\
\bottomrule
\end{tabular}
\end{table}

\subsection{Errors concentrate on a few grammatical words}
\label{sec:errors}

\begin{table}[t]
\centering\small
\caption{Tone accuracy (TAcc, \%) by word category, mean over three seeds. Last row: number
of aligned reference words (Direct).\protect\footnotemark}
\label{tab:breakdown}
\begin{tabular}{lrrrr}
\toprule
System & Fixed pattern & Contested & Phrase-medial & Phrase-final \\
\midrule
Direct & 98.5 & 88.1 & 94.8 & 94.1 \\
Explicit tone tokens & 98.5 & 87.7 & 94.6 & 94.1 \\
Toneless + lexical & 98.2 & 64.2 & 85.8 & 89.1 \\
Toneless + ByT5 & 98.4 & 86.0 & 93.9 & 94.3 \\
Toneless + source-aware ByT5 & 98.4 & 86.1 & 94.0 & 94.4 \\
\midrule
\textit{Aligned words (Direct)} & 8{,}379 & 4{,}503 & 12{,}191 & 715 \\
\bottomrule
\end{tabular}
\end{table}
\footnotetext{Only 24 aligned words are unseen in training, too few to report.}

Errors concentrate on contested words (Table~\ref{tab:breakdown}). A single contrast dominates: confusing \lmp{a} and \lmp{\`a}
accounts for 39\% of Direct's tone errors on contested words (205 of 527, seed~1), followed by
\lmp{laa}/\lmp{l\`aa}, \lmp{gee}/\lmp{g\`ee} and \lmp{ri\eng}/\lmp{r\`\i\eng}. Errors go both
ways: the model adds a mark the reference lacks (277 cases) about as often as it omits one
(201). These pairs are mostly distinct words rather than spelling variants. According to our native-speaker author and
the Limbum dictionary, plain \lmp{a} is the dependent pronoun \gl{they}, while \lmp{\`a} serves,
among other functions, as the third-person singular subject marker, a copula and a connective; the
test data fit this split (\lmp{Woow\`ee \textbf{a} ko' k\`una} \gl{They have gone up there} vs.\
\lmp{\eng gwe \textbf{\`a} koo y\`aa mb\ubar} \gl{A dog has caught my goat}). Because
\lmp{\`a} is multifunctional, however, the two forms are not always in contrast: our native-speaker
author judged \lmp{Ny\`{\ubar} a m boo rbuu ba nshe} \gl{God created the heaven and the earth}
equally correct with \lmp{a} or \lmp{\`a}. Part of the most frequent automatic error is
therefore an acceptable variant, consistent with the human evaluation (\S\ref{sec:human}). Other
contested pairs are clearly lexical:
\lmp{l\`aa} \gl{say} vs.\ \lmp{laa} \gl{cook}, \lmp{g\^ee} \gl{do} vs.\ \lmp{gee}
\gl{corner, side}, and \lmp{en\^{\eps}} \gl{this} vs.\ \lmp{en\eps} \gl{that}, so writing
\lmp{laa} where \lmp{l\`aa} is meant turns \gl{say} into \gl{cook}.

Overall accuracy does not differ by phrase position (94.1\% final vs.\ 94.8\% medial). A
few proper nouns do alternate between level and falling tone, e.g.\
\lmp{Tata}/\lmp{Tat\^a} \gl{Lord} and \lmp{Jus}/\lmp{J\^us}, which may reflect the
phrase-final falling tones described by \citet{gjersoe2019hybrid} or inconsistent spelling; the
Limbum dictionary confirms that the circumflex marks falling tone, but our data cannot separate
these explanations.

\subsection{Native-speaker evaluation}
\label{sec:human}
The native-speaker author found tone errors in 12 of the 100 Direct outputs (13 errors) and in 9 of
the 100 restoration outputs (10 errors). Most of them change the meaning: 10 of the 12 sentences
for Direct and 6 of the 9 for restoration. Overall preferences were balanced (restoration 51,
Direct 47, two ties; sign test $p=0.76$); because the two systems also differ in wording, these
preferences mainly reflect overall translation quality.

On the same 100 sentences, reference-based matching records 62 tone mismatches in 47 Direct
outputs and 72 in 56 restoration outputs. The annotator flagged 13 and 10, so only about one
automatic mismatch in five was judged an error. Some genuine errors may have been overlooked, but
the gap is consistent with the \lmp{a}/\lmp{\`a} case (\S\ref{sec:errors}): many mismatches with a
single reference are acceptable variants. Word-level tone accuracy against one reference therefore
overstates how often tone is wrong, and the small automatic difference between Direct and
restoration (62 vs.\ 72 mismatches) is not mirrored in the human counts (13 vs.\ 10). Tone errors
that a native speaker recognizes are infrequent, but most of them change what the sentence says.

\subsection{The legacy encoding distorts measurement more than models}

\begin{table}[t]
\centering\small
\caption{Direct models trained on legacy versus Unicode text, both scored on the same Unicode
references (legacy outputs converted with the same mapping). Mean $\pm$ s.d.\ over three
seeds. \chrfT$^{-q}$: quotation marks removed from hypotheses and references.}
\label{tab:legacy}
\begin{tabular}{lrrrr}
\toprule
Training text & \chrfT & \chrfT$^{-q}$ & TAcc & TAcc$_{con}$ \\
\midrule
Legacy encoding & 43.6\pmsd{0.3} & 43.9\pmsd{0.3} & 94.6\pmsd{0.1} & 87.5\pmsd{0.3} \\
Unicode & 44.3\pmsd{0.1} & 44.2\pmsd{0.2} & 94.7\pmsd{0.1} & 88.1\pmsd{0.2} \\
\bottomrule
\end{tabular}
\end{table}

Training on Unicode rather than legacy text gives a small gain in overall quality and no change
in tone accuracy (Table~\ref{tab:legacy}). The Unicode-trained Direct model gains 0.7 \chrfT, higher in all three seeds but clearly significant in only one ($p<0.001$;
$p=0.06$ and $0.08$ in the others). About half of the gain comes from punctuation: the legacy
models could not produce quotation marks, and with quotes removed the gap is 0.3 points.
Tone accuracy is unchanged (no significant difference in any seed): the model learns tone
equally well when it is written as stray letters after the vowel. Unicode training is also
more stable across seeds (s.d.\ 0.1 vs.\ 0.3). The legacy models do, however, produce invalid
text: across the six legacy runs (Direct and Explicit), 812 tone glyphs, about 135 per run,
are not attached to any vowel, a sequence that cannot be converted into valid Limbum.

The larger cost is to measurement. Scored in its own encoding, the legacy Direct model reaches
45.8 chrF++, 2.2 points more than the same outputs converted to Unicode, because each tone
glyph counts as an extra matching letter. In the legacy text, the acute accent marks the
vowels \eps\ and \ubar\ and appears to be a tone mark: an unconverted audit reports 48\% of
New Testament words as tone-marked and 57\% of running words as tone-ambiguous, and stripping
``tone'' merges \eps\ with \lmp{a}. Every tone-specific analysis on the unconverted corpus
would have been wrong, while the translation models themselves were only mildly affected.

\section{Discussion}

\paragraph{Measurement.}
For tonal languages written with diacritics, chrF++ should not be reported alone. Our
systems differ by less than one chrF++ point, while tone accuracy on contested words varies
from 64\% to 88\%, and a 12\% error rate on a third of running words moves chrF++ by under
two points. Word-level tone accuracy, reported with coverage and split by contested and
fixed-pattern words, makes these errors visible at no annotation cost, but it errs in the other
direction: against a single reference it counts acceptable variants as errors, and our
native-speaker author confirmed only about a fifth of its mismatches. Both measures are needed,
and for tonal languages a small native-speaker check is worth more than further automatic
precision.

\paragraph{Method.}
At this data scale, direct fine-tuning is the right choice. Explicit tone tokens add nothing
when the tokenizer already preserves tone marks, and translate-then-restore trades a small,
unreliable gain in overall chrF++ for a reliable loss in tone accuracy. Access to the English
source does not help the restorer, and the oracle ceiling shows that the remaining errors are
not a modelling failure that a better restorer could fix from written context alone. Progress
on contested words will more likely need more parallel data in context, or linguistic
information about the grammatical contrasts involved.

\paragraph{Relation to Yor\`ub\'a.}
Yor\`ub\'a restoration work starts from text in a language the pretrained model has seen, with
tens of thousands of sentences. In our unseen-language setting, NLLB-200 still learns
fixed-pattern tone almost perfectly from 7{,}000 pairs, which suggests that lexical tone is
cheap to acquire and that the hard part is the same in both settings: tone that depends on
grammatical context.

\paragraph{Written versus spoken tone.}
Our evaluation concerns tone as it is \emph{written}, not as it is spoken. The practical
orthography marks only low, falling and rising tone and leaves high and mid unmarked, so a
correctly spelled output still underdetermines pitch. Spoken Limbum tone also changes with
phrase position \citep{gjersoe2019hybrid}, and writers may mark lexical rather than surface
tone, or mark it inconsistently. Whether the written marks in this corpus match pronunciation
would require recordings of Limbum speakers aligned with the text; we leave this to future work,
which would also connect tone generation in text with Limbum speech technology.

\paragraph{Legacy encodings.}
Many African-language corpora derive from Bible translations typeset with legacy fonts.
Our results suggest that such encodings damage analysis and evaluation more than they damage
the models, which makes them easy to miss: the models train and score reasonably well. We
recommend checking character inventories per source and verifying against a known Unicode
text before any orthographic analysis.

\section{Conclusion}

A pretrained translation model fine-tuned on 7{,}000 sentence pairs learns to write tone in a
language it has never seen, almost perfectly for words with a fixed tone pattern and with
about one error in eight on contested words, the grammatical words whose tone depends on
context. These figures hold for the words the model translates correctly, about half of the
reference words; overall translation quality remains modest. Explicit tone tokens and translate-then-restore pipelines do not improve on direct
generation, and a standard metric barely registers the remaining errors. A legacy font
encoding hidden in the corpus would have invalidated every tone analysis while barely
affecting the models. Future work should test other model families, contrast unseen with seen tonal languages such
as Yor\`ub\'a, add independent native-speaker judgements, and compare written with spoken tone.
We release the converted corpus, code, model outputs and an annotation sheet for native-speaker
review.

\section*{Limitations}
The corpus is almost entirely biblical and has a single reference per sentence; the English
side (ASV) is archaic. Overall translation quality is modest (chrF++ 44--45), and tone accuracy
is measured only on the roughly half of reference words that systems translate correctly; our
tone results describe those words, not translation quality as a whole. The contested-word definition depends on spelling consistency in the
corpus, and part of what we count as contested may be inconsistent marking rather than
genuine contrast. Training was capped at 8 epochs and restorers were trained with one seed.
We study one unseen language and one pretrained model, NLLB-200; other model families, such as
models with much wider language coverage \citep{omt2026} or large language models used as
translators, may behave differently, and a seen tonal language would show how much of the
difficulty comes from Limbum being unseen. Automatic tone accuracy compares
spellings, not pronunciation, and cannot judge whether an alternative tone marking is
acceptable, which is why we include native-speaker evaluation. That evaluation
has a single annotator who is also an author and covers 100 sentences; blinding reduces but does
not remove possible bias, and we cannot report inter-annotator agreement. We could not recruit
further Limbum speakers for this study; independent judgements by several speakers are the most
important next step for confirming the human-evaluation findings. Finally, we evaluate written
tone only and do not test whether it matches spoken tone.

\section*{Ethical Considerations}
The source texts were produced for the Wimbum community, and the prior corpus release was
approved by an author of the dictionary \citep{ntohsi2026limbum}. Machine-generated Limbum
should not be deployed, for example for health or education materials, without review by
native speakers: tone errors that change meaning are exactly the failure that fluent-looking
output hides. The human evaluation was carried out by a native-speaker author, blind to system identity; no
external annotators were involved.

\acks{The authors received no specific funding for this work and declare no competing interests.

\noindent\textbf{AI Use Disclosure.} AI tools were used to assist with editing and improving the
clarity of the manuscript text and to create one figure. All AI-assisted content was reviewed and
verified by the authors.}

\newpage
\appendix
\section{Hyper-parameters}
\label{app:hparams}
\begin{table}[h]
\centering\small
\begin{tabular}{lll}
\toprule
 & NLLB-200-distilled-600M & ByT5-small restorer \\
\midrule
Optimizer & Adafactor, lr $10^{-4}$ & Adafactor, lr $5\times10^{-4}$ \\
Schedule & 200 warm-up, linear decay (floor 5\%) & same \\
Batch (per step $\times$ accum.) & $8\times2$ & $4\times4$ \\
Precision & fp16 (autocast) & fp32 \\
Max epochs / patience & 8 / 2 & 6 / 2 \\
Selection metric & dev \chrfT\ (greedy, 300 sent.) & dev TAcc \\
Max length & 128 src / 256 tgt tokens & 640 in / 448 out bytes \\
Decoding & beam 4 & greedy, 30-word chunks \\
New-tag init & mean(\texttt{swh}, \texttt{lug}, \texttt{lin}) & -- \\
Seeds & 3 & 1 \\
\bottomrule
\end{tabular}
\end{table}

\section{Reproducibility}
The companion notebook runs the encoding conversion, audit, all systems, the restorers,
bootstrap tests and table export end-to-end on a Colab T4. Every run writes its outputs and
is skipped when the notebook is re-run, so experiments can be resumed after a session ends.

\vskip 0.2in
\bibliography{references}

\begin{thebibliography}{23}
\providecommand{\natexlab}[1]{#1}
\providecommand{\url}[1]{\texttt{#1}}
\expandafter\ifx\csname urlstyle\endcsname\relax
  \providecommand{\doi}[1]{doi: #1}\else
  \providecommand{\doi}{doi: \begingroup \urlstyle{rm}\Url}\fi

\bibitem[Adelani et~al.(2021)Adelani, Ruiter, Alabi, Adebonojo, Ayeni, Adeyemi,
  Awokoya, and Espa{\~n}a-Bonet]{adelani2021menyo}
David~Ifeoluwa Adelani, Dana Ruiter, Jesujoba~O. Alabi, Damilola Adebonojo,
  Adesina Ayeni, Mofe Adeyemi, Ayodele~Esther Awokoya, and Cristina
  Espa{\~n}a-Bonet.
\newblock The effect of domain and diacritics in {Y}or{\`u}b{\'a}--{E}nglish
  neural machine translation.
\newblock In \emph{Proceedings of Machine Translation Summit XVIII: Research
  Track}, pages 61--75, 2021.

\bibitem[Adelani et~al.(2022)Adelani, Alabi, Fan, Kreutzer, Shen, Reid, Ruiter,
  Klakow, Nabende, Chang, et~al.]{adelani2022few}
David~Ifeoluwa Adelani, Jesujoba~Oluwadara Alabi, Angela Fan, Julia Kreutzer,
  Xiaoyu Shen, Machel Reid, Dana Ruiter, Dietrich Klakow, Peter Nabende, Ernie
  Chang, et~al.
\newblock A few thousand translations go a long way! leveraging pre-trained
  models for {A}frican news translation.
\newblock In \emph{Proceedings of the 2022 Conference of the North American
  Chapter of the Association for Computational Linguistics: Human Language
  Technologies}, pages 3053--3070, 2022.

\bibitem[Ager(2025)]{omniglotlimbum}
Simon Ager.
\newblock Limbum.
\newblock Omniglot: The Online Encyclopedia of Writing Systems and Languages.
  \url{https://www.omniglot.com/writing/limbum.htm}, 2025.
\newblock Last modified 8 October 2025; sample text (Luke 11:2--4) reproduced
  from Bible.com. Accessed October 2026.

\bibitem[Dale(2023)]{dale2023nllb}
David Dale.
\newblock How to fine-tune a {NLLB}-200 model for translating a new language.
\newblock Blog post, Medium.
  \url{https://cointegrated.medium.com/how-to-fine-tune-a-nllb-200-model-for-translating-a-new-language-a37fc706b865},
  2023.
\newblock Accessed October 2026.

\bibitem[Fiore(1987)]{fiore1987}
Lynne Fiore.
\newblock \emph{A Phonology of {Limbum}}.
\newblock SIL Cameroon, 1987.

\bibitem[Fransen(1995)]{fransen1995}
Margo Astrid~Eleonora Fransen.
\newblock \emph{A Grammar of {Limbum}: A {G}rassfields {B}antu Language Spoken
  in the {N}orth-{W}est Province of {C}ameroon}.
\newblock PhD thesis, Vrije Universiteit Amsterdam, 1995.

\bibitem[Gjers{\o}e et~al.(2019)Gjers{\o}e, Nformi, and
  Paschen]{gjersoe2019hybrid}
Siri Gjers{\o}e, Jude Nformi, and Ludger Paschen.
\newblock Hybrid falling tones in {Limbum}.
\newblock In Emily Clem, Peter Jenks, and Hannah Sande, editors, \emph{Theory
  and Description in {A}frican Linguistics: Selected Papers from the 47th
  Annual Conference on {A}frican Linguistics}, pages 95--118. Language Science
  Press, Berlin, 2019.
\newblock \doi{10.5281/zenodo.3367130}.

\bibitem[Johnson et~al.(2017)Johnson, Schuster, Le, Krikun, Wu, Chen, Thorat,
  Vi{\'e}gas, Wattenberg, Corrado, Hughes, and Dean]{johnson2017google}
Melvin Johnson, Mike Schuster, Quoc~V. Le, Maxim Krikun, Yonghui Wu, Zhifeng
  Chen, Nikhil Thorat, Fernanda Vi{\'e}gas, Martin Wattenberg, Greg Corrado,
  Macduff Hughes, and Jeffrey Dean.
\newblock {Google}'s multilingual neural machine translation system: Enabling
  zero-shot translation.
\newblock \emph{Transactions of the Association for Computational Linguistics},
  5:\penalty0 339--351, 2017.

\bibitem[Koehn(2004)]{koehn2004statistical}
Philipp Koehn.
\newblock Statistical significance tests for machine translation evaluation.
\newblock In \emph{Proceedings of the 2004 Conference on Empirical Methods in
  Natural Language Processing}, pages 388--395, 2004.

\bibitem[Kudo and Richardson(2018)]{kudo2018sentencepiece}
Taku Kudo and John Richardson.
\newblock {SentencePiece}: A simple and language independent subword tokenizer
  and detokenizer for neural text processing.
\newblock In \emph{Proceedings of the 2018 Conference on Empirical Methods in
  Natural Language Processing: System Demonstrations}, pages 66--71, 2018.

\bibitem[N{\'a}plava et~al.(2018)N{\'a}plava, Straka, Stra{\v{n}}{\'a}k, and
  Haji{\v{c}}]{naplava2018diacritics}
Jakub N{\'a}plava, Milan Straka, Pavel Stra{\v{n}}{\'a}k, and Jan Haji{\v{c}}.
\newblock Diacritics restoration using neural networks.
\newblock In \emph{Proceedings of the Eleventh International Conference on
  Language Resources and Evaluation ({LREC} 2018)}, 2018.

\bibitem[Neubig and Hu(2018)]{neubig2018rapid}
Graham Neubig and Junjie Hu.
\newblock Rapid adaptation of neural machine translation to new languages.
\newblock In \emph{Proceedings of the 2018 Conference on Empirical Methods in
  Natural Language Processing}, pages 875--880, 2018.

\bibitem[{NLLB Team} et~al.(2022){NLLB Team}, Costa-juss{\`a}, Cross,
  {\c{C}}elebi, Elbayad, Heafield, Heffernan, Kalbassi, Lam, Licht, Maillard,
  et~al.]{nllb2022}
{NLLB Team}, Marta~R. Costa-juss{\`a}, James Cross, Onur {\c{C}}elebi, Maha
  Elbayad, Kenneth Heafield, Kevin Heffernan, Elahe Kalbassi, Janice Lam,
  Daniel Licht, Jean Maillard, et~al.
\newblock No language left behind: Scaling human-centered machine translation.
\newblock \emph{arXiv preprint arXiv:2207.04672}, 2022.

\bibitem[Ntohsi et~al.(2026)Ntohsi, Tuyishimire, Kafesu, Ogore, Babalola, and
  Ankeli]{ntohsi2026limbum}
Samiratu Ntohsi, Neza~David Tuyishimire, Anesu Kafesu, Marvin Ogore,
  Samuel~Oluwajunwonlo Babalola, and Oche Ankeli.
\newblock Embedding initialization for unseen low-resource languages in
  multilingual {NMT}: A case study on {Limbum}--{English} translation.
\newblock \emph{arXiv preprint arXiv:2608.07629}, 2026.

\bibitem[Olawole et~al.(2024)Olawole, Alabi, Sakpere, and
  Adelani]{olawole2024yad}
Akindele~Michael Olawole, Jesujoba~O. Alabi, Aderonke~Busayo Sakpere, and
  David~I. Adelani.
\newblock {YAD}: Leveraging {T5} for improved automatic diacritization of
  {Y}or{\`u}b{\'a} text.
\newblock \emph{arXiv preprint arXiv:2412.20218}, 2024.

\bibitem[{Omnilingual MT Team} et~al.(2026){Omnilingual MT Team}, Alastruey,
  Bafna, Caciolai, Heffernan, Kozhevnikov, Ropers, S{\'a}nchez, Saint-James,
  Tsiamas, Cao, Cheng, Chuang, Duquenne, Duppenthaler, Ekberg, Gao,
  Huguet~Cabot, Janeiro, Maillard, Mejia~Gonzalez, Schwenk, Toledo, Turkatenko,
  Ventayol-Boada, Moritz, Mourachko, Parimi, Williamson, Yates, Dale, and
  Costa-juss{\`a}]{omt2026}
{Omnilingual MT Team}, Belen Alastruey, Niyati Bafna, Andrea Caciolai, Kevin
  Heffernan, Artyom Kozhevnikov, Christophe Ropers, Eduardo S{\'a}nchez,
  Charles-Eric Saint-James, Ioannis Tsiamas, Xiang Cao, Chierh Cheng, Joe
  Chuang, Paul-Ambroise Duquenne, Mark Duppenthaler, Nate Ekberg, Cynthia Gao,
  Pere-Llu{\'i}s Huguet~Cabot, Jo{\~a}o~Maria Janeiro, Jean Maillard, Gabriel
  Mejia~Gonzalez, Holger Schwenk, Edan Toledo, Arina Turkatenko, Albert
  Ventayol-Boada, Rashel Moritz, Alexandre Mourachko, Surya Parimi, Mary
  Williamson, Shireen Yates, David Dale, and Marta~R. Costa-juss{\`a}.
\newblock Omnilingual {MT}: Machine translation for 1,600 languages.
\newblock arXiv:2603.16309, 2026.
\newblock URL \url{https://arxiv.org/abs/2603.16309}.

\bibitem[Orife(2018)]{orife2018attentive}
Iroro Orife.
\newblock Attentive sequence-to-sequence learning for diacritic restoration of
  {Y}or{\`u}b{\'a} language text.
\newblock In \emph{Proceedings of Interspeech 2018}, pages 2848--2852, 2018.

\bibitem[Orife et~al.(2020)Orife, Adelani, Fasubaa, Williamson, Oyewusi, Wahab,
  and Tubosun]{orife2020improving}
Iroro Orife, David~I. Adelani, Timi Fasubaa, Victor Williamson, Wuraola~Fisayo
  Oyewusi, Olamilekan Wahab, and Kola Tubosun.
\newblock Improving {Y}or{\`u}b{\'a} diacritic restoration, 2020.
\newblock arXiv preprint.

\bibitem[Popovi{\'c}(2017)]{popovic2017chrf}
Maja Popovi{\'c}.
\newblock {chrF++}: Words helping character n-grams.
\newblock In \emph{Proceedings of the Second Conference on Machine
  Translation}, pages 612--618, 2017.

\bibitem[Post(2018)]{post2018call}
Matt Post.
\newblock A call for clarity in reporting {BLEU} scores.
\newblock In \emph{Proceedings of the Third Conference on Machine Translation:
  Research Papers}, pages 186--191, 2018.

\bibitem[Shazeer and Stern(2018)]{shazeer2018adafactor}
Noam Shazeer and Mitchell Stern.
\newblock Adafactor: Adaptive learning rates with sublinear memory cost.
\newblock In \emph{Proceedings of the 35th International Conference on Machine
  Learning}, volume~80 of \emph{Proceedings of Machine Learning Research},
  pages 4596--4604, 2018.

\bibitem[Stankevi{\v{c}}ius et~al.(2022)Stankevi{\v{c}}ius,
  Luko{\v{s}}evi{\v{c}}ius, Kapo{\v{c}}i{\=u}t{\.e}-Dzikien{\.e},
  Brie{\v{d}}ien{\.e}, and Krilavi{\v{c}}ius]{stankevicius2022byt5}
Lukas Stankevi{\v{c}}ius, Mantas Luko{\v{s}}evi{\v{c}}ius, Jurgita
  Kapo{\v{c}}i{\=u}t{\.e}-Dzikien{\.e}, Monika Brie{\v{d}}ien{\.e}, and Tomas
  Krilavi{\v{c}}ius.
\newblock Correcting diacritics and typos with a {ByT5} transformer model.
\newblock \emph{CoRR}, abs/2201.13242, 2022.

\bibitem[Xue et~al.(2022)Xue, Barua, Constant, Al-Rfou, Narang, Kale, Roberts,
  and Raffel]{xue2022byt5}
Linting Xue, Aditya Barua, Noah Constant, Rami Al-Rfou, Sharan Narang, Mihir
  Kale, Adam Roberts, and Colin Raffel.
\newblock {ByT5}: Towards a token-free future with pre-trained byte-to-byte
  models.
\newblock \emph{Transactions of the Association for Computational Linguistics},
  10:\penalty0 291--306, 2022.

\end{thebibliography}

\end{document}